\documentclass{article}
\usepackage{spconf,amsmath,graphicx,hyperref}
\usepackage{cite}
\usepackage{amsmath,amssymb,amsfonts}
\usepackage{algorithmic}
\usepackage{graphicx}
\usepackage{textcomp}
\usepackage{multirow}
\usepackage[table]{xcolor}
\usepackage{xcolor}
\usepackage{url}
\usepackage{enumitem}

\title{PROST-LLM: Progressively Enhancing the Speech-to-Speech Translation Capability in LLMs}
\name{Jing Xu$^{\dagger}$, Jiaqi Wang$^{\dagger}$, Daxin Tan$^{\star}$, Xiao Chen$^{\star}$}
\address{$^{\dagger}$The Chinese University of Hong Kong\\
$^{\star}$Huawei Artificial Intelligence Laboratory (Leibniz)}
\begin{document}
\ninept
\maketitle
\begin{abstract}
Although Large Language Models (LLMs) excel in many tasks, their application to Speech-to-Speech Translation (S2ST) is underexplored and hindered by data scarcity. To bridge this gap, we propose PROST-LLM (\textbf{PRO}gressive \textbf{S}peech-to-speech \textbf{T}ranslation) to enhance the S2ST capabilities in LLMs progressively. 
First, we fine-tune the LLMs with the CVSS corpus, employing designed tri-task learning and chain of modality methods to boost the initial performance. 
Then, leveraging the fine-tuned model, we generate preference pairs through self-sampling and back-translation without human evaluation. 
Finally, these preference pairs are used for preference optimization to enhance the model's S2ST capability further. 
Extensive experiments confirm the effectiveness of our proposed PROST-LLM in improving the S2ST capability of LLMs.
\end{abstract}
\begin{keywords}
Speech-to-speech translation, preference optimization, back-translation, large language model
\end{keywords}
\section{Introduction}
\label{sec:intro}
Speech-to-Speech Translation (S2ST) bridges communication gaps between speakers of different languages. While Large Language Models (LLMs) have shown remarkable potential in diverse domains \cite{gpt4,yang2024qwen2,llama3.2,team2024gemini}, their application in S2ST remains underexplored and hindered by the scarcity of paired S2ST data. Moreover, although preference optimization has proven effective for enhancing the performance of LLMs in Natural Language Processing (NLP) tasks, its potential for improving the S2ST performance of LLMs has not been studied.

Conventional S2ST systems typically adopted cascaded pipelines \cite{janus,wahlster2013verbmobil,atr}, integrating Automatic Speech Recognition (ASR), Machine Translation (MT), and Text-to-Speech (TTS) synthesis. 
The first end-to-end approach Translatron \cite{translatron} and its improved version, Translatron2 \cite{translatotron2}, employed a multi-objective task to train a sequence-to-sequence model. Recent studies\cite{uwspeech,textless,huang2022transpeech} have explored to use discrete speech representations as an intermediate representation of end-to-end S2ST, reducing computation, inference latency, and error accumulation. However, these methods often involve complex architectures and scale poorly. In contrast, LLM-based S2ST methods offer a simpler architecture and the ability to exploit multi-task and multi-modality correlations, but enabling LLMs to perform S2ST effectively with limited S2ST paired data remains an open challenge.

Preference optimization (PO) is a common method for aligning the outputs of LLMs with human preferences, enabling performance enhancement even with limited data. Traditional preference optimization methods like RLHF \cite{rlhf} align LLMs with user intent through human feedback, depending on costly human feedback. Recent offline approaches like DPO \cite{dpo} and SimPO \cite{simpo} have been introduced to reduce this requirement but rely on high-quality preference pairs. For example, research \cite{zhangadaptiveqr} used marginal probabilities of answers as the reward to construct paired data for DPO training. However, how to construct the required preference data pairs for S2ST tasks remains under-investigated.

In machine translation, back-translation translates text from the source language to the target language and then back to the source language. Findings from \cite{brislin1970back} suggest that translation quality can be predicted via comparisons between original and back-translated text. Subsequent studies \cite{edunov2018understanding, sugiyama2019data} adopted back-translation with monolingual data to perform data augmentation, achieving higher translation accuracy. Inspired by this, we introduce back-translation to automatically assess the quality of model-generated translations, and use the resulting assessments to construct preference data pairs for PO in S2ST tasks, eliminating the need for human evaluation.

In this paper, we present PROST-LLM (\textbf{PRO}gressive-\textbf{S}peech-to-speech \textbf{T}ranslation) to progressively enhance S2ST capability in LLMs. First, we fine-tune the pre-trained LLMs on the CVSS corpus with our designed tri-task learning (ASR, speech translation (S2T), S2ST) or chain of modality (text prediction preceding speech prediction) strategy to strengthen task understanding. The fine-tuned models are then used for self-sampling and back-translation evaluation to automatically construct preference pairs, which are subsequently used in preference optimization, eliminating the need for large scale paired data or costly human annotations. Our contributions include: 
\begin{itemize}[leftmargin=3ex]
    \item We present the first framework to equip LLMs with S2ST capability via tri-task and chain-of-modality learning.
    \item As far as we know, we are the first to apply preference optimization to further enhance the S2ST capability of LLMs via back-translation-driven preference pairs construction.
    \item We utilize monolingual speech corpora to construct preference pairs, thereby reducing the reliance of LLMs on paired data for S2ST tasks.
    \item Experiments demonstrate that PROST-LLM reduces the BLEU gap between the end-to-end and cascaded S2ST system to 1.2, validating the effectiveness of PROST-LLM.
\end{itemize}
\begin{figure*}[ht]
\label{fig: prost_llm framework}
\centering
\begin{minipage}[b]{0.55\linewidth}
    \centering
\centerline{\includegraphics[width=\linewidth]{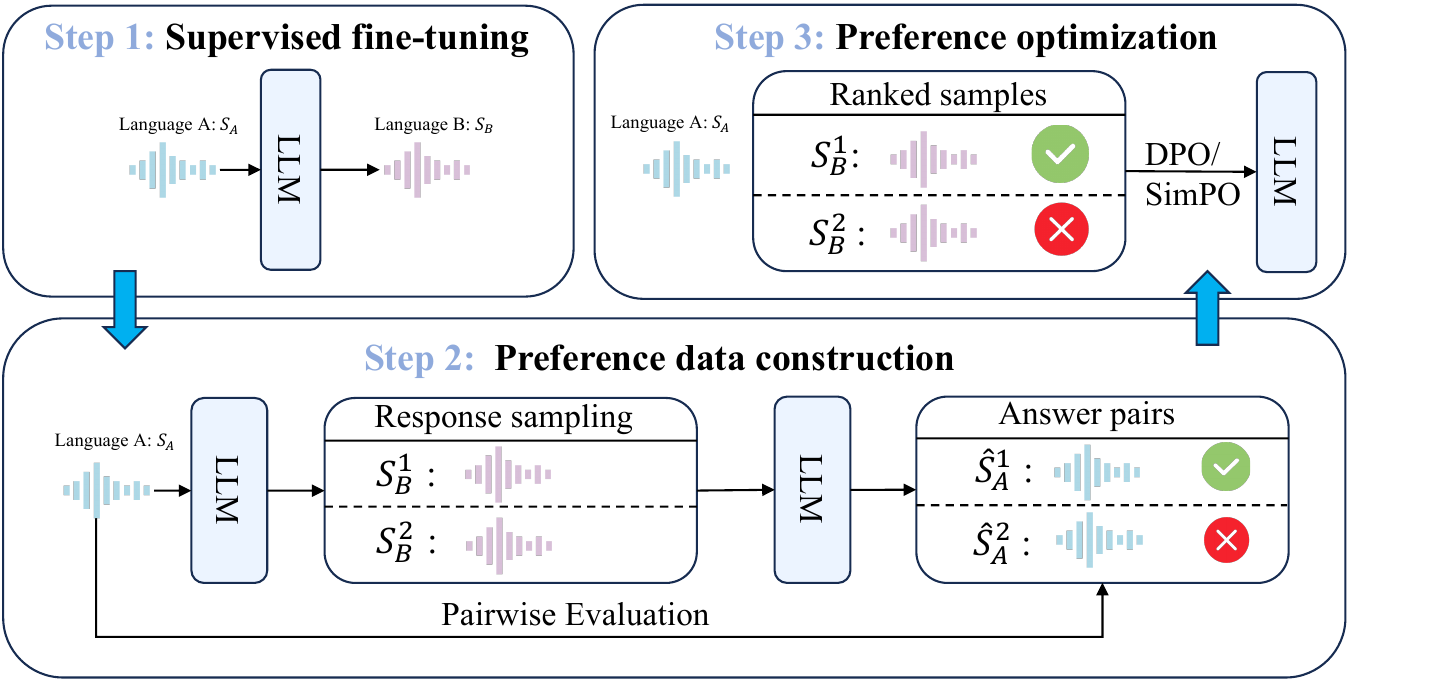}}
  \centerline{(a)}\medskip
\end{minipage}
\hfill
\begin{minipage}[b]{0.445\linewidth}
\centering
  \centerline{\includegraphics[width=\linewidth]{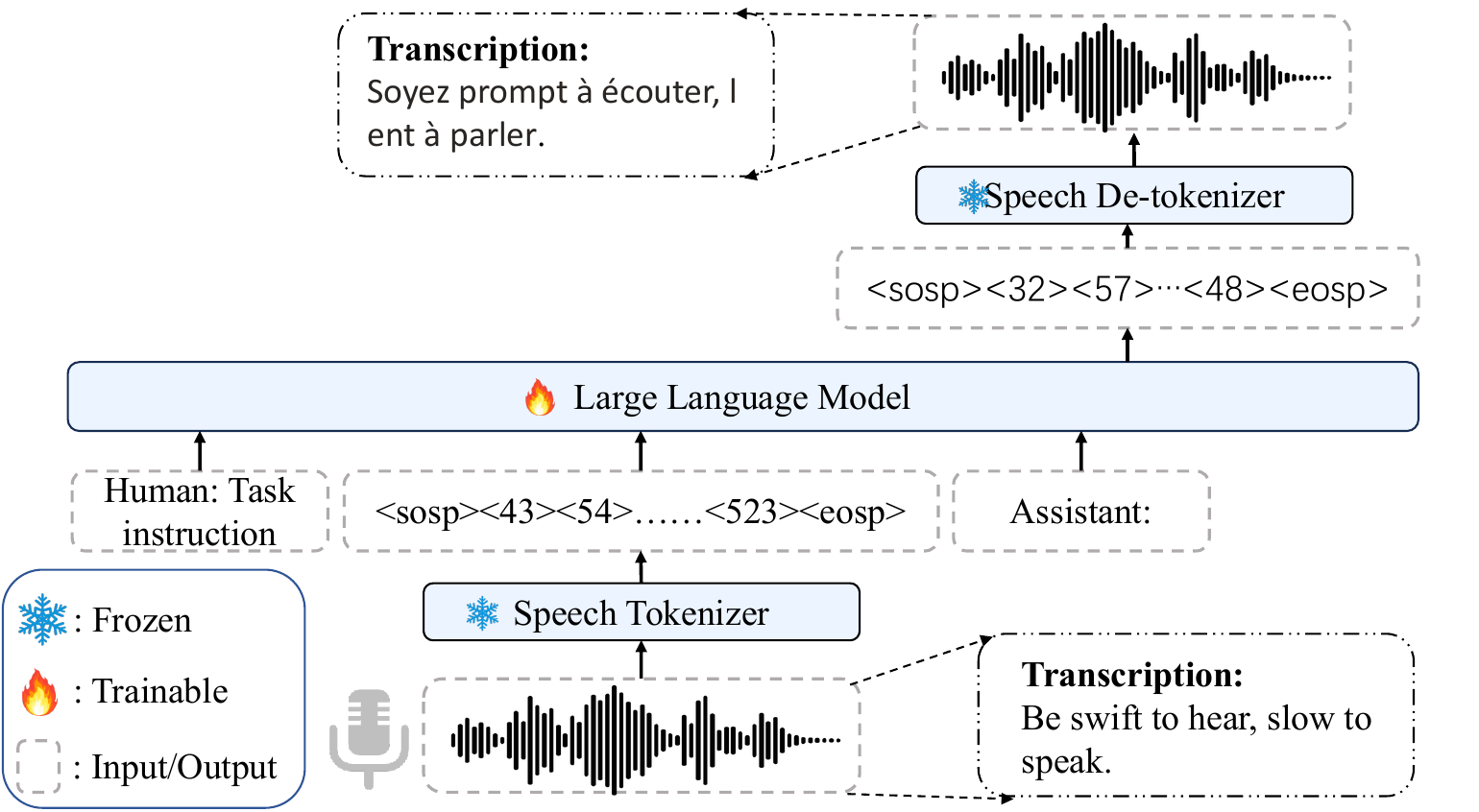}}
  \centerline{(b)}\medskip
\end{minipage}
\vspace{-2em}
\caption{(a) Our PROST-LLM training system: (i) Step 1: Supervised fine-tuning (SFT) the LLM. (ii) Step 2: Based on the SFT LLM, we construct preference data pairs (\textit{e.g.,} $(S_A, S_B^1, S_B^2)$) by comparing back-translated answer pairs $\hat{S}_A^1, \hat{S}_A^2$ with the ground truth $S_A$. (iii) Step 3: Preference optimizing the SFT LLM using the constructed preference data pairs. (b) The architecture of PROST-LLM in step 1. }
\end{figure*}


\section{Methodology}
We introduce PROST-LLM, a framework to improve S2ST ability in LLMs via supervised fine-tuning (SFT) and PO. Section~\ref{subsec:overview} provides an overview of the framework. Section \ref{subsec:sft} details the supervised fine-tuning process with tri-task learning and chain of modality strategies. Section \ref{subsec:backtranslation} explains how preference data are constructed. Section \ref{subsec:po} describes the preference optimization procedure.

\subsection{Overview}
\label{subsec:overview}
As shown in Figure \ref{fig: prost_llm framework} (a), PROST-LLM proceeds in three main steps:

\textbf{Step 1: Supervised fine-tuning.} The model is fine-tuned on the CVSS corpus to acquire initial S2ST ability.

\textbf{Step 2: Preference data construction.} The fine-tuned model generates candidate translations, which are back-translated into the source language. Comparing the back-translations with original speech enables automatic preference pairs generation without human evaluation (See Section \ref{subsec:backtranslation}).

\textbf{Step 3: Preference optimization.} Preference optimization is applied to the fine-tuned model using the constructed preference data pairs to further improve S2ST performance.

\subsection{Supervised fine-tuning}
\label{subsec:sft}

As shown in Figure~\ref{fig: prost_llm framework} (b), PROST-LLM consists of a speech tokenizer, an LLM backbone and a speech de-tokenizer. Speech inputs are discretized via clustering on speech representations extracted using SSL. These units are added to the LLM’s vocabulary and embedding matrix, enabling end-to-end fine-tuning on S2ST tasks. The model learns to translate speech units across languages, and the de-tokenizer reconstructs waveforms from generated units.

Given the scarcity of S2ST corpora, we design two strategies to enhance the model's understanding of speech modality and S2ST tasks:
\begin{itemize}[leftmargin=3ex]
    \item\textbf{Tri-task learning.} The S2ST corpus is converted into parallel ASR and S2T data, enabling joint training on ASR, S2T, and S2ST with task-specific instructions. This encourages cross-task knowledge transfer between ASR, S2T and S2ST.
    \item\textbf{Chain of modality.} To simplify direct speech-to-speech generation, the model is instructed to first generate target text, then produce target speech based on the text within a single forward pass. This sequential output structure facilitates cross-modal learning and improves generation stability.
\end{itemize}

\subsection{Preference data construction}
\label{subsec:backtranslation}

For each source language speech $S_A$, the fine-tuned model $M_{SFT}$ generates two translation candidates $\left\{S_B^i \right\}_{i=1}^2$, which are then back-translated into source language speech $\left\{\hat{S}_A^i \right\}_{i=1}^2$ using the same model. Supervised fine-tuning ensures these candidates are of reasonable quality, while back-translation enables automatic assessment without costly human labeling. 
We assess translation quality using multiple metrics: \textbf{\textit{(i)}} acoustic measures (Mel-Cepstral Distortion, MCD, for spectral similarity), \textbf{\textit{(ii)}} text-based metrics (Word Error Rate, WER, for transcription accuracy), and \textbf{\textit{(iii)}} translation metrics (BLEU \cite{bleu} for n-gram overlap, METEOR \cite{meteor} for precision, recall, and semantic alignment).

After comparison, each candidate $\left\{S_B^i \right\}_{i=1}^2$ receives a score, with low-is-better metrics such as WER, MCD inverted so that higher scores consistently indicate better quality. For each candidate pair, we designate the preferred $S_B^p$ and rejected $S_B^r$ if $e_p-e_r>\delta$, where $\delta>0$ is a margin hyper-parameter. This yields preference data pairs $\left\{(S_A, S_B^p, S_B^r) \right\}$ without human annotation, while also enabling the use of monolingual speech corpora.
\subsection{Preference optimization}
\label{subsec:po}
\begin{table*}
\centering
\vspace{-0.6em}
\caption{Comparison of different systems on CVSS corpus. Bold marks the best.}
\label{tab:main}
\resizebox{0.95\linewidth}{!}{
\scalebox{0.8}{
\begin{tabular}{lcccccccccc}
\hline
\multicolumn{1}{c}{\multirow{3}{*}{System}}          & \multicolumn{5}{c}{CVSS-C}  & \multicolumn{5}{c}{CVSS-T}\\
\cline{2-11} & \multicolumn{2}{c}{S2T(BLEU$\uparrow$)}  & \multicolumn{2}{c}{S2ST(BLEU$\uparrow$)}& UTMOS$\uparrow$&\multicolumn{2}{c}{S2T(BLEU$\uparrow$)}  & \multicolumn{2}{c}{S2ST(BLEU$\uparrow$)}&UTMOS$\uparrow$ \\
\cline{2-11}    & en2fra & fra2en  & en2fra & fra2en&fra2en & en2fra & fra2en & en2fra & fra2en&fra2en\\ 
\hline \hline
\multicolumn{11}{l}{\cellcolor[HTML]{ECF4FF}\textbf{Cascaded System}}\\
\hspace{0.8em}S2T+TTS  & 29.27 &     24.40   &  28.27 &22.82&$3.04\pm0.24$& 26.83 &   24.54   &  25.37     &    22.91&$3.08\pm0.27$\\\hline\hline

\multicolumn{11}{l}{\cellcolor[HTML]{ECF4FF}\textbf{End-to-end system (Before DPO)}}\\
\hspace{0.8em}Vanilla     & - &      -    & 14.65 & 13.99&$3.71\pm0.23$& - &    -  &  12.53     & 13.20 &$3.64\pm0.30$     \\
\hspace{0.8em}Tri-Task Learning  & 23.61 &  20.29   & 18.68& 15.15& $3.65\pm0.24$&21.99 &  17.35   &  14.14        &  16.46 &$3.63\pm0.21$ \\
\hspace{0.8em}Chain of Modality  & 29.21  &  21.96          & 24.20 & 20.79&$3.68\pm0.29$& 23.20 &   18.42   &     18.61 &   16.93 &$3.60\pm0.31$  \\\hline\hline

\multicolumn{11}{l}{\cellcolor[HTML]{ECF4FF}\textbf{End-to-end system (After DPO)}}\\
\multicolumn{11}{l}{\cellcolor[HTML]{FFF0F5}\hspace{0.4em}\textbf{Back-translation evaluation metric - METEOR}}\\
\hspace{0.8em}Vanilla & - &   -    &   19.33 &  15.25& \textbf{3.74\:$\pm$\:0.22} &- &- &16.95 & 15.92 &$3.64\pm0.25$ \\
\hspace{0.8em}Tri-Task Learning & 24.65 &    21.23   & 20.71 & 16.72 &$3.68\pm0.23$&22.03 &20.94 & 17.11 & 17.20&$3.66\pm0.23$\\  
\hspace{0.8em}Chain of Modality & 29.94 &  22.78    &  25.04 & 21.59&$3.66\pm0.29$ &25.84 &20.61 &22.24 & \textbf{19.49}&$3.61\pm0.23$ \\\hline
\multicolumn{11}{l}{\cellcolor[HTML]{FFF0F5}\hspace{0.4em}\textbf{Back-translation evaluation metric - BLEU}}\\
\hspace{0.8em}Vanilla & - &  -    & 17.93  & 15.25&$3.72\pm0.26$ & - &    -      &   16.75   &   15.34 &$3.61\pm0.28$ \\ 
\hspace{0.8em}Tri-Task Learning & 24.65 &    21.15  &  20.96 &  16.68&$3.66\pm0.22$& 22.20 &   20.86      &    16.95  &  17.24 &\textbf{3.67\:$\pm$\:0.22}  \\ 
\hspace{0.8em}Chain of Modality &  \textbf{29.97} &   \textbf{23.04} & \textbf{25.12}  & \textbf{21.78}&$3.70\pm0.29$ & \textbf{26.13} &     \textbf{20.75}     &   \textbf{22.54}   &  19.47 &$3.61\pm0.29$  \\ 
\hline 
 \hline 
\end{tabular}}
}

\end{table*}
Using the constructed preference pairs $\left\{(S_A, S_B^p, S_B^r) \right\}$, we further fine-tune $M_{SFT}$ with Direct Preference Optimization (DPO), which maximizes the margin between preferred and rejected translations:
\begin{equation}
\begin{aligned}
    L_{DPO}  = & -\mathbb{E}_{(S_A, S_B^p, S_B^r) \sim D} \left[ \log \sigma \left(\beta \log \frac{M_{\theta}(S_B^p \mid S_A)}{M_{SFT}(S_B^p \mid S_A)}  \right. \right.  \\
        &\left. \left. - \beta \log \frac{M_{\theta}(S_B^r \mid S_A)}{M_{SFT}(S_B^r \mid S_A)} \right) \right],
\end{aligned}
\end{equation}
where  $M_{\theta}$ is initialized from $M_{SFT}$, $\sigma$ is the sigmoid function and $\beta$ is a scaling hyper-parameter. This step enhances the model’s ability to distinguish and generate high-quality speech translations, improving cross-modal reasoning and overall S2ST performance.

\section{Experimental setup}
\subsection{Training configuration}
We focus on English-French translation. Speech representations are extracted with mHuBERT \cite{textless}, discretized via K-means, and reconstructed with unit HiFi-GAN vocoders trained separately on English and French. All pretrained components are from the official repository\footnote{\url{https://github.com/facebookresearch/fairseq/blob/main/examples/speech_to_speech/docs/textless_s2st_real_data.md}}. For bidirectional translation, input and output languages are simply reversed. 
The LLM backbone is LLaMA 3.2-3B\cite{llama3.2}, while Whisper-large-v3\cite{whisper} transcribes both source and back-translated speech. Unless otherwise specified, preference optimization is trained with 5,000 samples per direction. 

For training, we adopt two stages: (1) Supervised fine-tuning, where the LLM is fully fine-tuned for 4 epochs with batch size 64, learning rate $1\times 10^{-4}$; (2) Preference optimization, where LoRA with rank 8 is applied to all linear layers except the LM head. This stage is trained for 2 epochs with batch size 32, and learning rate $2\times 10^{-5}$. We set the margin hyper-parameter $\delta$ for preference pair selection to $0.1$. A pair is retained only if the score gap exceeds $\delta$, ensuring more reliable preference supervision. BLEU is used as the default evaluation metric.
\subsection{Datasets}

We use the French-English subset of CVSS\cite{cvss}, which includes sentence-level parallel S2ST pairs translated into English from 21 languages, derived from Commonvoice\cite{commonvoice} and CoVoST 2\cite{covost}. 
Translated speech is synthesized using two TTS systems, resulting in two dataset variants: CVSS-C (174 hours, single speaker) and CVSS-T (192.7 hours, multiple speakers). 
Each sample in CVSS is formatted as $(S_{sr}, T_{sr}, S_{tgt}, T_{tgt})$, where $S_{sr}$ and $T_{sr}$ denote source language speech and text, while $S_{tgt}$ and $T_{tgt}$ represent target language speech and text. 
For tri-task learning, $(S_{sr}, T_{sr})$ and $(S_{tgt}, T_{tgt})$ are used for ASR task, $(S_{sr}, T_{tgt})$ and $(S_{tgt}, T_{sr})$ are used for S2T. 
\subsection{Evaluation metrics}
Translation quality is measured using BLEU scores between ASR-transcribed outputs and reference text. Since ASR may introduce errors, the resulting BLEU score serves as a lower bound for translation quality. Additionally, speech naturalness is evaluated with UTMOS\cite{utmos}, an automatic Mean Opinion Score (MOS) predictor that correlates well with human evaluations and perceptions.
\section{Experimental results}
In all tables, "en" denotes English, "fra" denotes French, and a notation such as "A2B" indicates translation from source language A to target language B (e.g., en2fra means English to French translation, fra2en means French to English translation).
\subsection{Comparison systems}
To evaluate the effectiveness of PROST-LLM, we compare it against a cascaded S2ST system composed of an S2T model and a robust TTS system\footnote{\url{https://github.com/myshell-ai/MeloTTS}}. The S2T component shares the same architecture and training dataset as PROST-LLM.

In addition, we implement three supervised fine-tuning variants: 
\begin{itemize}[leftmargin=3ex]
    \item\textbf{Vanilla:} An LLM-based S2ST system without any specialized designs.
    \item\textbf{Tri-task learning:} Jointly training on ASR, S2T and S2ST tasks to promote cross-task knowledge transfer.
    \item\textbf{Chain of modality:} Sequentially generating intermediate target text before speech units to bridge modalities without tri-task learning.
\end{itemize}
By contrasting tri-task learning and chain of modality with the vanilla setting, we can directly assess how each design contributes to improving S2ST performance beyond simple SFT.

\subsection{Main results}

Table~\ref{tab:main} shows that both tri-task learning and chain of modality outperform the vanilla baseline, confirming that: (1) LLMs benefit from learning correlated tasks, and (2) generating intermediate text facilitates direct S2ST learning.  Notably, chain of modality yields greater gains, highlighting its effectiveness in bridging speech-text modality gaps. In terms of speech quality, UTMOS scores show that PROST-LLM produces more natural speech than the cascaded baseline, likely due to stronger speaker consistency.

Preference optimization provides additional performance boosts, even with only 5k preference pairs per direction. On CVSS-C, BLEU improves from 0.84 (en2fra under chain of modality setting using METEOR) to 4.68 (en2fra under vanilla setting using METEOR), while on CVSS-T, BLEU improves from 0.86 (fra2en under tri-task learning setting using METEOR) to 4.42 (en2fra under vanilla setting using METEOR). Interestingly, S2T performance under tri-task learning also improves after PO, confirming our hypothesis (Sec.~\ref{subsec:po}) that PO enhances LLM comprehension of speech and its alignment with text. Overall, consistent gains are observed across training strategies and evaluation metrics. Moreover, the BLEU gap between PROST-LLM and cascaded system narrows dramatically: from 14.38 to 3.15 (en2fra) and from 8.83 to 1.04 (fra2en), demonstrating both the generalization ability and  effectiveness of PROST-LLM.

\subsection{Monolingual vs. S2ST preference data}
\label{subsec: monolingual}
\begin{table}[!h]
\centering
\vspace{-1.2em}
\caption{Monolingual vs. S2ST preference data}
\label{tab: monolingual}
\resizebox{0.9\linewidth}{!}{
\begin{tabular}{lcccc}
\hline
\multicolumn{1}{c}{\multirow{2}{*}{System}} & \multicolumn{2}{c}{S2T(BLEU$\uparrow$)}  & \multicolumn{2}{c}{S2ST(BLEU$\uparrow$)}\\
\cline{2-5}    & en2fra & fra2en  & en2fra & fra2en\\ 
\hline \hline

\hspace{0.8em}Cascaded system & 26.83 &   24.54   &     25.37 &   22.91  \\\hline \hline
 \multicolumn{5}{l}{\cellcolor[HTML]{ECF4FF}\textbf{End-to-end system (After DPO)}}\\
\hspace{0.8em}Chain of Modality(S2ST) & 26.13 &     20.75     &   22.54   &  19.47 \\
\hspace{0.8em}Chain of Modality(Mono) & \textbf{27.83}  &  \textbf{22.97}     &  \textbf{23.72}  &  \textbf{21.71}   \\\hline 
 \hline 
\end{tabular}
}
\end{table}
As discussed in Section \ref{subsec:backtranslation}, back-translation enables the use of monolingual speech corpora for preference optimization. We therefore compare models trained on preference pairs derived from monolingual corpora versus those from the paired S2ST corpus. Specifically, we use the English subset of Commonvoice 4.0 and French subset of Commonvoice 19.0. For robust evaluation, we select the chain of modality model trained on CVSS-T, as CVSS-C’s single-speaker data may cause out-of-distribution degradation. Moreover, the chain of modality model demonstrates stronger understanding of the S2ST task by explicitly bridging speech and text modalities, making it more resilient to domain shifts

Table~\ref{tab: monolingual} shows that using preference data derived from monolingual corpora yields even better S2ST performance, reducing BLEU score gap between cascaded and end-to-end systems to 1.65 (en2fra) and 1.2 (fra2en). This observation highlights the advantage of monolingual corpora utility in reducing paired S2ST data reliance.

\subsection{Influence of back-translation metrics}
\begin{table}[!ht]
\centering
\vspace{-1.4em}
\caption{Comparison of different back-translation metrics}
\label{tab: evaluation method}
\resizebox{0.9\linewidth}{!}{
\begin{tabular}{lcccc}
\hline
\multicolumn{1}{c}{\multirow{2}{*}{System}}  & \multicolumn{2}{c}{S2T(BLEU$\uparrow$)}  & \multicolumn{2}{c}{S2ST(BLEU$\uparrow$)}\\
\cline{2-5}    & en2fra & fra2en  & en2fra & fra2en\\ 
\hline \hline



\multicolumn{5}{l}{\hspace{0.4em}\textbf{Back-translation evaluation metric - WER}}\\
\hspace{0.8em}Vanilla & - &   -    &  18.34  &  15.23   \\
\hspace{0.8em}Tri-Task Learning & 24.36 &   21.04    & 20.19 & 16.30  \\  

\hspace{0.8em}Chain of Modality & \textbf{30.05} &   \textbf{22.90}    &  25.10  & \textbf{21.69}  \\\hline 
 
\multicolumn{5}{l}{\hspace{0.4em}\textbf{Back-translation evaluation metric - MCD}}\\
\hspace{0.8em}Vanilla &  -&     -  & 18.43   &   14.70   \\
\hspace{0.8em}Tri-Task Learning  & 24.24 &   20.54   & 20.09 & 16.18 \\  

\hspace{0.8em}Chain of Modality& 29.79 &    22.42   &  \textbf{25.34}  &  21.26     \\\hline 
 \hline 
\end{tabular}
}
\vspace{-0.6em}
\end{table}
 As shown in Table~\ref{tab: evaluation method} and Table \ref{tab:main}, translation quality improves consistently across all back-translation evaluation metrics, verifying robustness of our approach.. Translation-based metrics generally outperform acoustic and text based metrics, while MCD is particularly effective for English-to-French translation. This may be due to consistent speaker identity in the output speech units, making spectral comparisons more reliable.

\subsection{Scalability across other PO algorithms and settings}

We also explore Simple Preference Optimization (SimPO) as the preference optimization method to assess the scalability of PROST-LLM. The results in Table~\ref{tab: optimization method} demonstrate that PROST-LLM is agnostic to the specific preference optimization algorithm. 
 \begin{table}[!ht]
\centering
\vspace{-1.4em}
\caption{Comparison of different PO methods on CVSS-C}
\label{tab: optimization method}
\resizebox{0.82\linewidth}{!}{
\begin{tabular}{lcccc}
\hline
\multicolumn{1}{c}{\multirow{2}{*}{System}}          & \multicolumn{2}{c}{S2T(BLEU$\uparrow$)}  & \multicolumn{2}{c}{S2ST(BLEU$\uparrow$)}\\
\cline{2-5}    & en2fra & fra2en  & en2fra & fra2en\\ 
\hline \hline




\multicolumn{5}{l}{\hspace{0.4em}\textbf{DPO}}\\
\hspace{0.8em}Vanilla & - &   -    &  17.93  &   15.25  \\
\hspace{0.8em}Tri-Task Learning & 24.65 &   21.15    & 20.96 & 16.68  \\  

\hspace{0.8em}Chain of Modality & \textbf{29.97}  &  \textbf{23.04}     &  25.12  &  \textbf{21.78}  \\\hline 
 
\multicolumn{5}{l}{\hspace{0.4em}\textbf{SimPO}}\\
\hspace{0.8em}Vanilla & - &   -    &  15.79  &  14.23   \\
\hspace{0.8em}Tri-Task Learning & 24.20 &   20.63    & 20.02 & 16.08  \\  

\hspace{0.8em}Chain of Modality & 29.72  &  22.36     &  \textbf{25.14}  &  21.20  \\\hline 
 \hline 
\end{tabular}
}
\vspace{-1.3em}
\end{table}

We further conduct an ablation study on training sample size and iteration count in the preference optimization step. As shown in Table~\ref{tab: training number}, increasing training sample size improves performance. Notably, even with just 2,500 samples, performance enhancements are observed, demonstrating the efficiency of our PROST-LLM systems. For models that have not yet reached optimal performance with the current data size (as seen in vanilla and tri-task learning settings), additional iterations further boost performance. In contrast, if the model already shows satisfactory performance after a single iteration (as in the chain of modality setting with 5,000 samples), additional iterations offer no extra gains, suggesting convergence to optimal performance. In such cases, incorporating monolingual speech corpora can further enhance performance as discussed in Section \ref{subsec: monolingual}. 

Additionally, when comparing a single iteration with 5,000 samples to two iterations with 2,500 samples each, we find that single iteration training achieves comparable performance to two-iteration training. Thus, single iteration training is recommended due to its lower computation and time costs.
\begin{table}[!ht]
\centering
\vspace{-1.4em}
\caption{Ablation on training sample size and iterative count on CVSS-C}
\label{tab: training number}
\resizebox{\linewidth}{!}{
\setlength{\tabcolsep}{2pt}
\begin{tabular}{lcccccccc}
\hline
\multicolumn{1}{c}{\multirow{3}{*}{System}} & \multicolumn{4}{c}{1 iteration} & \multicolumn{4}{c}{2 iterations} \\
\cline{2-9}& \multicolumn{2}{c}{S2T(BLEU$\uparrow$)}  & \multicolumn{2}{c}{S2ST(BLEU$\uparrow$)}   & \multicolumn{2}{c}{S2T(BLEU$\uparrow$)}  & \multicolumn{2}{c}{S2ST(BLEU$\uparrow$)}   \\
\cline{2-9}    & en2fra & fra2en  & en2fra & fra2en& en2fra & fra2en  & en2fra & fra2en\\ 
\hline \hline

\multicolumn{9}{l}{\hspace{0.4em}\textbf{5,000 samples for each iteration}}\\
\hspace{0.8em}Vanilla & - &   -    &  17.93  &   15.25 &-  &    -   &  18.85  &   15.43 \\
\hspace{0.8em}Tri-Task Learning & 24.65 &   21.15    & 20.96 & 16.68 & 24.74 &  21.34    & 21.03   &  16.83 \\  

\hspace{0.8em}Chain of Modality & \textbf{29.97}  &  \textbf{23.04}     &  \textbf{25.12}  &  \textbf{21.78} & \textbf{30.05}  &  \textbf{23.05}     &  25.08  &  \textbf{21.75}  \\ \hline
\multicolumn{9}{l}{\hspace{0.4em}\textbf{2,500 samples for each iteration}}\\
\hspace{0.8em}Vanilla & - &   -    &  17.02  & 15.00  & - &   -    & 18.43   &  15.27 \\  
\hspace{0.8em}Tri-Task Learning & 24.28 &   20.75    &  20.27  &  16.39  & 24.27 &   20.80    & 20.44 & 16.41   \\
\hspace{0.8em}Chain of Modality & 29.56  &   22.29    &   24.97 &  21.13  & 29.79  &  22.56     &  \textbf{25.16}  &  21.33 \\\hline 
 \hline 


\end{tabular}
}
\vspace{-0.6em}
\end{table}
\section{Conclusion}
We propose PROST-LLM, a framework for progressively enhancing speech-to-speech translation (S2ST)  capabilities in LLMs. Starting with supervised fine-tuning on the CVSS corpus, we introduce tri-task learning and chain-of-modality to boost initial S2ST ability in LLMs. Using the fine-tuned model, we generate preference data pairs via self-sampling and back-translation, enabling preference optimization to refine translation quality without costly human labels. Experiments show that PROST-LLM significantly improves S2ST quality, reduces dependence on paired S2ST data, and remains robust to different back-translation metrics and PO algorithms. Future work will extend PROST-LLM to more languages, more LLM backbones and larger monolingual speech resources.

\small
\bibliographystyle{IEEEbib}
\bibliography{strings,refs}

\end{document}